\documentclass[journal,onecolumn,11pt]{IEEEtran}

\usepackage[T1]{fontenc}
\usepackage[utf8]{inputenc}
\usepackage{amsmath,amssymb}
\usepackage{graphicx}
\usepackage{booktabs}
\usepackage{array}
\usepackage{multirow}
\usepackage{xcolor}
\usepackage{tikz}
\usetikzlibrary{arrows.meta,positioning,calc}
\usepackage{cite}
\usepackage{listings}
\usepackage{xurl}
\usepackage{xspace}
\usepackage[hidelinks]{hyperref}

\definecolor{accent}{RGB}{38,110,210}
\definecolor{muted}{RGB}{190,190,180}

\newcolumntype{L}[1]{>{\raggedright\arraybackslash}p{#1}}

\newcommand{\g}{\,g\xspace}

\begin{document}

\title{Single-Image Food Mass Estimators Do Not Transfer Their Gram Scale Across Datasets}

\author{Yongsheng~Lian%
\thanks{Y. Lian is with the University of Louisville, Louisville, KY, USA (e-mail: yongsheng.lian@louisville.edu).}%
\thanks{This work received no external funding. The author declares no competing interests.}}

\maketitle

\begin{abstract}
Estimating food mass from a photograph is central to image-based dietary assessment, but whether trained models transfer to new cameras and foods is unclear. We fine-tuned DINOv2-base on 2,253 weighed Nutrition5k dishes to predict mass from one RGB photograph and compared it with a zero-shot vision-language model, Qwen3-VL-32B, on Nutrition5k and three external datasets: ASA24 studio portions, ECUSTFD whole foods and Food Portion Benchmark meal trays. In domain, the fine-tuned model's mean absolute error was 24.8\g, less than half of Qwen's. On external datasets it did not beat Qwen. For a food it had never seen, it ranked portions by size, but no better than a count of food pixels. Its gram scale was wrong in two ways: it changed with how much of the scene the photograph showed, and it was off by a different factor for each food. Training on other foods or trays from the tested dataset cut the external error by 38 to 65\%. DINOv2-large, a larger backbone trained on the same dishes, did not reduce the external error. One labeled photograph of the same food cut the model's ASA24 error to 28.3\g, better than that photograph achieved alone. Prediction intervals that held the true mass for 90\% of Nutrition5k dishes held it for at most half of external photographs. Nutrition5k often photographed the same plate several times, so a random test split shared near-identical photographs with training and overstated accuracy. Single-image mass estimators therefore need labeled photographs from the setting where they are used.
\end{abstract}


\section{Introduction}

\IEEEPARstart{A}{ccurate} dietary assessment underpins nutrition research and clinical care, yet food records and 24-hour recalls are burdensome and prone to misreporting. Image-based assessment could reduce this burden, but estimating nutrients from a photograph requires identifying the foods and estimating how much of each is present. Quantity is often the harder step, because a single photograph does not fix volume, food density varies and mixed dishes hide components~\cite{thames2021nutrition5k,lo2020review}.

Supervised image regressors trained on weighed food can estimate dish mass well: networks trained on Nutrition5k reach errors near 20\g~\cite{thames2021nutrition5k,shao2023rgbd,han2023dpf}. They were, however, tested only on held-out photographs from Nutrition5k itself, taken with the same overhead camera, in the same cafeteria and of similar dishes. Whether they stay accurate on photographs taken with another camera, from another distance or of other foods, as a deployed system would receive, has not been tested. Vision-language models (VLMs), also called multimodal large language models, estimate portions without being trained on any one dataset and can be grounded in food databases~\cite{yan2025dietai24,coburn2026openknead}, but whether they are more or less accurate than a trained model on such new photographs is also unknown. Since one photograph cannot fix absolute size, a model that is accurate in domain must have learned the size cues of its training setting, and it is unknown whether those cues hold elsewhere.

We fine-tuned DINOv2 to predict a dish's mass from its photograph alone, trained on weighed Nutrition5k dishes, and compared it with a zero-shot VLM, Qwen3-VL-32B, on Nutrition5k and on three external datasets that differ in camera, framing, viewing angle, foods and label source. We separate two kinds of accuracy. \emph{Ordering} is the ability to rank portions correctly, for example to see that one serving is twice another. The \emph{gram scale} is the ability to put the right number of grams on a portion, which also requires knowing the camera's framing and the food's density. A model can keep the first and lose the second. The study is a retrospective benchmark on public images; it does not evaluate a clinical dietary assessment tool.

The study makes three main observations:
\begin{enumerate}
  \item \textbf{Accurate in domain, no better than a zero-shot VLM elsewhere.} The fine-tuned model's mean absolute error (MAE) on Nutrition5k was less than half of Qwen's, but on the three external datasets it was no more accurate than Qwen.
  \item \textbf{The gram scale fails in two ways, and what transfers is trivial.} The scale shifted with framing and camera angle, and it was off by a different factor for each unseen food. The model still ranked the portions of a new food correctly, but a count of food pixels did as well.
  \item \textbf{Only data from the same setting restores the scale.} Training photographs from the same dataset, of foods or trays not in its test set, cut the external error by 38 to 65\%, whereas a larger backbone did not; for a known food, one labeled photograph cut the error further.
\end{enumerate}
Two further observations support the same conclusion: prediction intervals calibrated in domain failed out of domain, and a random Nutrition5k test split overstated accuracy because it shares near-duplicate plates with the training set.

\section{Related Work}

\subsection{Image-Based Mass Estimation}

Early systems paired food classification with geometric or learned volume estimation, using reference objects, multiple views or depth sensors to recover scale~\cite{meyers2015im2calories,lo2020review}. Nutrition5k provides about 5,000 cafeteria dishes weighed ingredient by ingredient, with overhead RGB-D images and side-angle video~\cite{thames2021nutrition5k}. Supervised networks trained on it reached mass MAEs of 40.4\g for the original RGB baseline~\cite{thames2021nutrition5k}, 27.5\g for Swin-Nutrition~\cite{shao2022swin} as re-implemented on the overhead photographs~\cite{han2023dpf}, 21.6\g for an RGB-D fusion network that needs a depth camera at prediction time~\cite{shao2023rgbd} and 21.2\g for DPF-Nutrition, which predicts depth from the RGB image~\cite{han2023dpf}. Vision Transformers fine-tuned on side-angle video frames reached 61 to 92\g~\cite{andrade2025pretraining}. None of these networks was evaluated outside Nutrition5k, and Swin-Nutrition, the RGB-D network and DPF-Nutrition provided no usable code or weights, so their transfer to other cameras and foods is unknown. Few public datasets pair food photographs with weights: ECUSTFD shows weighed whole foods beside a coin~\cite{liang2017ecustfd}, the Food Portion Benchmark (FPB) shows laboratory-weighed meals from several viewpoints~\cite{fpb2025}, and the NCI ASA24 image library shows foods at standard portions linked to USDA weights~\cite{nci_asa24_images}.

\subsection{VLM-Based Dietary Assessment}

VLM pipelines for dietary assessment typically let the VLM identify the foods and then look up nutrients in a food database. DietAI24 links GPT-4 Vision to the USDA Food and Nutrient Database for Dietary Studies (FNDDS) through retrieval-augmented generation: the VLM chooses FNDDS food codes and a standard portion such as ``1 cup'', and the weight in grams is read from FNDDS rather than estimated~\cite{yan2025dietai24,usda_fndds}. On 1,000 Nutrition5k images it reported a food-weight MAE of 45.1\g. Open-KNEAD, a training-free pipeline built on open VLMs, instead lets the VLM estimate grams directly and uses FNDDS to code the foods and supply their nutrient composition, not their weight~\cite{coburn2026openknead}. Other evaluations give the model the ingredient list, which cut ChatGPT-5's energy error from 123 to 53\,kcal in one study~\cite{rodriguez2025chatgpt5} and improved most VLMs' nutrient estimates on Nutrition5k in another~\cite{vlm2026comparative}; such results do not measure estimation from a photograph alone.

Liao and Li combine the two families~\cite{liao2026geometry}: a commercial multimodal LLM names and locates each food and supplies a density range, and a small head on frozen DINOv2 features estimates each food's mass. Trained and evaluated separately within Nutrition5k, FPB and NutritionVerse-Real, it reduced per-food error by 33 to 41\% relative to the LLM alone; transfer of one model across datasets was not tested.

\subsection{Positioning of This Study}
\label{sec:positioning}

Previously published supervised models, namely the baseline network released with the Nutrition5k dataset~\cite{thames2021nutrition5k}, Swin-Nutrition~\cite{shao2022swin}, RGB-D Nutrition~\cite{shao2023rgbd}, DPF-Nutrition~\cite{han2023dpf} and the Vision Transformers of Andrade et al.~\cite{andrade2025pretraining}, were trained and tested only on Nutrition5k. Liao and Li~\cite{liao2026geometry} used three datasets, Nutrition5k, FPB and NutritionVerse-Real, but trained a separate model on each dataset's training split and tested it only on that dataset's test split. None of these models was tested on a dataset it was not trained on. The key contribution of this work is to test whether such an estimate transfers: the same fixed models, a fine-tuned DINOv2 and a zero-shot VLM, were scored once on three external test sets that differ in framing, camera angle, foods and label source. We separate how well a model orders portions from whether it gives the right number of grams, and measure which data restore the gram scale when it fails. Our estimate differs from Liao and Li's in kind: they estimate each food's mass with the help of the LLM's food names and boxes, whereas we estimate the dish's total mass from the photograph alone, so we do not compare the results numerically; their one dish-level Nutrition5k figure is listed for context in Appendix~\ref{sec:decision}.

\section{Materials and Methods}

\subsection{Study Design}

This retrospective study used four public datasets (Table~\ref{tab:datasets}). The unit of analysis was one photograph and its total food mass. Models were trained and tuned on the training sets, and on Nutrition5k also on the validation set; the test sets were used only for testing, and each external test set was scored once for each model declared in advance. Every photograph received a prediction. Only public datasets were analyzed~\cite{thames2021nutrition5k,nci_asa24_images,usda_fndds,liang2017ecustfd,fpb2025}; no human participants were involved, and ethical approval was not required.

External results fall into three settings, which support different claims. In \emph{cross-dataset transfer}, a model is scored on a dataset none of whose photographs it was trained on. In \emph{same-dataset training}, a model trained with the ASA24 or FPB training photographs is scored on that dataset's test set, whose foods or trays never appear in training. In \emph{known-food adaptation}, the model is corrected with labeled photographs of the test foods themselves (Section~\ref{sec:fewshot}), so it needs the identity of the food as well as the photograph.

\begin{table*}[t]
\centering
\caption{Datasets and their roles. CV, cross-validation.}
\label{tab:datasets}
\resizebox{\textwidth}{!}{%
\begin{tabular}{lrL{4.6cm}L{3.0cm}L{4.6cm}}
\toprule
Dataset (part) & Images & Content & Mass label & Role \\
\midrule
Nutrition5k training set & 2,253 & Overhead cafeteria plates & Weighed & Training (5-fold CV) \\
Nutrition5k validation & 502 & Same & Weighed & Method selection \\
Nutrition5k internal test & 498 & Same & Weighed & Supporting only; near-duplicates of training (Sec.~\ref{sec:siblings}) \\
ASA24 training set & 5,099 (6,329) & NCI studio portion photographs & FNDDS standard weights & Training; CV grouped by food \\
ASA24 test & 1,000 & Same; 162 foods, none in the training set & FNDDS standard weights & External test; once per pre-declared model \\
ECUSTFD & 1,429 & Top views of 146 whole foods (19 types), coin for scale & Weighed per object & External test; never trained on \\
FPB training set & 9,190 & Meal trays and items, five viewpoints & Laboratory-weighed per item & Training (Nutrition5k+ASA24+FPB model); CV grouped by tray \\
FPB test set (official validation split) & 2,351 (2,034) & Same & Same & External test; primary subset (2,034) excludes trays seen in training \\
\bottomrule
\end{tabular}}
\end{table*}

\subsection{Datasets}
\label{sec:n5ksets}
\label{sec:external}

\paragraph{Nutrition5k} Nutrition5k contains about 5,000 cafeteria dishes whose ingredients were weighed one by one~\cite{thames2021nutrition5k}; we use the overhead RGB photographs and, as a training target only, the depth maps. Its official split has 2,758 training and 507 test dishes with overhead photographs. We formed three disjoint sets (Appendix~\ref{sec:n5kresults}):
\begin{itemize}
  \item \textbf{Training set (2,253 dishes):} the official training split minus our internal test set and three dishes without a photograph.
  \item \textbf{Validation set (502 dishes):} the official test split minus five dishes with faulty records. We used it to choose methods and settings, so its errors are slightly optimistic.
  \item \textbf{Internal test set (498 dishes):} drawn from the official training split, stratified by dish weight and ingredient count (Appendix~\ref{app:sampling}). It was scored six times, each time for models fixed beforehand, so we treat it as supporting evidence only.
\end{itemize}
Nutrition5k plates were often photographed repeatedly as foods were added. Because our internal test set was drawn at random, 84\% of its dishes, against 6\% of validation dishes, have such a near-duplicate in the training set (Section~\ref{sec:siblings}). We therefore take the validation set as the main in-domain estimate.

\paragraph{ASA24} The NCI ASA24 Portion Size Image Database contains 17,135 studio photographs, each showing a food at a stated portion, such as ``Iceberg lettuce, raw, 3 cups''~\cite{nci_asa24_images}. The released images are $160\times160$-pixel cut-outs, which we placed on a white background. The database links each photograph to an FNDDS food code, a standard portion and a multiplier~\cite{usda_fndds}, and we computed its label as the FNDDS weight of that portion times the multiplier, $55\times3=165$\g for the lettuce. The label is thus a standard weight, not a weighing: half of a small and a quarter of a large wedge of angel food cake both receive 16.975\g (Appendix~\ref{sec:labelaudit}). ASA24 often links one photograph to several similar foods: a photograph of a quarter cup of chocolate drink powder, for example, is linked to cocoa beverage powder, milk beverage powder and malted milk powder, which FNDDS weighs at 44.0, 36.4 and 19.2\g for that portion. The true mass of such a photograph is uncertain, so we labeled it with the median of its foods' weights only when the heaviest was at most 25\% above the lightest, and otherwise left it unlabeled, as for the powder. In all, 6,469 photographs were labeled (Appendix~\ref{sec:labelaudit}). ASA24 photographs each food at several portion sizes on the same plate, so each food has a series of photographs. We split the data by food, so that no food appears in both training and test. The \textbf{test set} has 1,000 overhead photographs: the complete series of 162 foods, drawn at random with a fixed seed before any model was run. The \textbf{training set} has 5,099 photographs of 719 other foods; we removed every photograph that showed a test food or was linked to one of its FNDDS codes. The \textbf{expanded training set}, used for our best ASA24 model, has 6,329 photographs of 845 foods; it also keeps photographs whose linked foods mostly, rather than all, agree in weight (Appendix~\ref{sec:labelaudit}). ASA24 has no separate validation set, so we chose models by 5-fold cross-validation on the training set, with all photographs of a food in the same fold; we call these errors the \emph{ASA24 validation} results.

\paragraph{ECUSTFD} We used 1,429 top-view photographs of 146 weighed whole foods of 19 types, each photographed several times beside a 25\,mm coin~\cite{liang2017ecustfd}. ECUSTFD was used only for testing.

\paragraph{FPB} The Food Portion Benchmark (FPB) contains about 14,000 laboratory photographs of 138 Central Asian foods, taken from above and from side angles, with the weight of every item measured~\cite{fpb2025}. A photograph's label is the sum of its item weights. We used the official training split, 9,190 photographs, for training, and the official validation split, 2,351 photographs, as our \textbf{test set}; labels of the official test split are not public. We removed 317 test photographs whose tray, identified by its list of items and weights, also appears in training, leaving 2,034. Training photographs are mostly single items, with a median of 204\g, whereas test photographs are mostly trays of several items, with a median of 456\g. Photographs were reduced to at most 1,024 pixels on the longer side; most are portrait $768\times1{,}024$ smartphone images.

\subsection{Fine-Tuned DINOv2}
\label{sec:training}

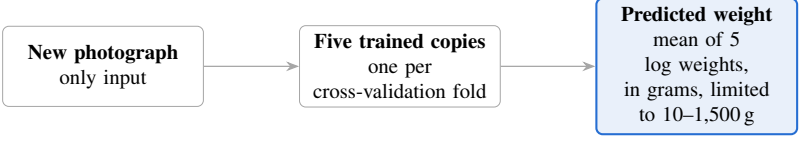
\begin{figure*}[t]
\centering
\resizebox{\textwidth}{!}{%
\begin{tikzpicture}[
  font=\small,
  box/.style={draw=gray!70, rounded corners=3pt, align=center, minimum height=1.35cm, text width=3.1cm, inner sep=4pt},
  train/.style={box, dashed},
  outbox/.style={box, draw=accent, line width=1pt, fill=accent!10},
  arr/.style={-{Stealth[length=2.2mm]}, draw=gray!70, line width=0.6pt},
  darr/.style={arr, dashed},
  lbl/.style={font=\footnotesize, text=gray!80!black},
  panel/.style={font=\bfseries, anchor=west}
]
\node[panel] at (-1.8,2.7) {a\quad Training};
\node[box] (img) at (0,0) {\textbf{Photograph}\\$448\times336$, flips and\\brightness jitter,\\no cropping};
\node[box] (enc) at (4,0) {\textbf{DINOv2}\\pretrained, all layers\\fine-tuned};
\node[box] (feat) at (8,0) {\textbf{Image features}\\{[CLS]} token +\\mean patch token};
\node[box] (head) at (12,0) {\textbf{Mass head}\\512-unit layer\\$\to$ 1 output,\\log weight};
\node[box] (loss) at (16,0) {\textbf{Loss}\\error on log weight\\+ depth error};
\node[train] (dhead) at (12,-2.4) {\textbf{Depth head}\\1 linear unit,\\training only, then\\discarded};
\node[train] (dtarget) at (19.6,-2.4) {\textbf{Depth target}\\height-sum proxy\\from depth camera};
\node[box] (mtarget) at (19.6,0) {\textbf{Measured weight}\\log grams};
\draw[arr] (img) -- (enc);
\draw[arr] (enc) -- (feat);
\draw[arr] (feat) -- (head);
\draw[arr] (head) -- (loss);
\draw[arr] (mtarget) -- (loss);
\draw[darr] (feat.south) |- (dhead.west);
\draw[darr] (dhead.east) -| ($(loss.south)+(-0.45,0)$);
\draw[darr] (dtarget.west) -| ($(loss.south)+(0.45,0)$);
\draw[arr] (loss.north) -- ++(0,0.75) -| node[lbl, above, pos=0.25] {update all weights; repeat for 15 passes} (enc.north);
\node[panel] at (-1.8,-4.3) {b\quad Prediction};
\node[box] (new) at (0,-5.7) {\textbf{New photograph}\\only input};
\node[box] (copies) at (5,-5.7) {\textbf{Five trained copies}\\one per\\cross-validation fold};
\node[outbox] (out) at (10,-5.7) {\textbf{Predicted weight}\\mean of 5 log weights,\\in grams, limited\\to 10--1,500\,g};
\draw[arr] (new) -- (copies);
\draw[arr] (copies) -- (out);
\end{tikzpicture}}
\caption{Fine-tuning DINOv2 to predict food weight. (a) During training, DINOv2 and a small head predict the logarithm of the dish weight from the photograph. For Nutrition5k dishes, a second head predicts a depth-derived height-sum proxy; dashed elements are used only during training. Both errors are combined and all weights are updated. (b) For prediction, only the photograph is needed: it passes through the five copies trained in 5-fold cross-validation, and their log weights are averaged.}
\label{fig:training}
\end{figure*}

DINOv2 is a vision transformer pretrained by self-supervision on about 142 million images~\cite{oquab2024dinov2}. We fine-tune the whole model to predict the logarithm of a dish's mass from its photograph (Fig.~\ref{fig:training}). Photographs are resized to $448\times336$ pixels, flipped and varied in brightness, but never cropped, because the apparent size of the food carries information about its amount. For Nutrition5k dishes, a second head predicts a height-sum proxy from the depth camera. Depth thus serves as privileged information, which helps training but is not needed at prediction time~\cite{vapnik2009lupi}: the head is discarded after training, so the model needs only the photograph. Five copies are trained in 5-fold cross-validation, and a new photograph's mass is the average of their predictions~\cite{isensee2021nnunet}. Appendix~\ref{app:training} gives every setting.

We name the models by their training data. \emph{Nutrition5k-only} models were trained on Nutrition5k; the main one is DINOv2-base. \emph{Two-domain} models were trained on Nutrition5k and ASA24; four were compared (Appendix~\ref{sec:drivers}), and the main one is DINOv2-large trained with the expanded ASA24 training set. The \emph{three-domain} model, DINOv2-base, was trained on Nutrition5k, ASA24 and FPB. All used only the training photographs of these datasets. Tables and figures name the models by their training data, Nutrition5k+ASA24 and Nutrition5k+ASA24+FPB. We added datasets one at a time, rather than training a single model on all of them, because each step answers a separate question. The Nutrition5k-only model has seen none of the external datasets, so it measures transfer. Adding ASA24 tests whether photographs from a new setting, of foods other than those tested, restore accuracy in that setting, while FPB and ECUSTFD stay unseen and show whether that gain carries over to other settings. Adding FPB repeats the test for meal trays. A single model trained on all datasets could not separate these effects, nor tell a gain from same-dataset photographs apart from a gain from more or more varied training data; for that, controls with a larger backbone and with FPB alone isolate the effect of same-dataset photographs (Section~\ref{sec:multidomain}). ECUSTFD was never used in training, so that every model could be tested on a dataset it had never seen; with 146 objects of 19 food types, it is also too small to split into useful training and test sets. Each recipe used 15 passes through its training data, so adding photographs also added optimizer steps. To check what drives the results, we also trained the model with a frozen backbone and without the depth target, a fine-tuned ResNet-50~\cite{he2016resnet}, DINOv2-large on Nutrition5k alone and DINOv2-base on FPB alone, and repeated three Nutrition5k-only variants with three further seeds. The seed, backbone and FPB-only controls were run after the test sets had been scored and were evaluated only on data that are not test sets (Appendix~\ref{sec:drivers}).

\subsection{Zero-Shot VLMs}
\label{sec:vlm}

We used Qwen3-VL-32B-Instruct~\cite{bai2025qwen3vl}, served with vLLM~\cite{kwon2023vllm} at temperature 0.01. In one call per photograph, it received the photograph without any food names, identified each food item itself and returned a gram estimate for each item; the dish estimate was their sum. The prompts are given in Appendix~\ref{app:prompts}. Responses without a usable estimate after one retry, at most 1.4\% in any run, received the training-set median weight. To check that the results are not specific to Qwen, we ran a second, low-cost VLM, OpenAI's \texttt{gpt-6-luna}, with the same prompt and with reasoning switched off. Neither VLM was trained on any of our data. We also tested two Qwen variants, one given the ingredient names and one corrected by its errors on similar training dishes; their results are summarized in Section~\ref{sec:zeroshot}.

\subsection{Outcomes and Statistical Analysis}
\label{sec:stats}

The primary outcome was mass MAE. Two further measures separate ordering from gram scale. The Spearman correlation between predicted and true mass measures how well the predictions rank the photographs by mass; it is 1 when the ranking is exactly right, whatever the gram values. We computed it across all photographs of a test set and, on ASA24, within each food's series of portions. The scale ratio, the geometric mean of predicted over true mass, measures whether a model systematically over- or underestimates; it is 1 when there is no such bias, and 0.5 when the model typically predicts half the true mass. We compare two methods by the difference in their MAEs, first method minus second, so a negative difference means that the first method erred less. To judge whether a difference could be due to the particular photographs in a test set, we used a paired bootstrap: we redrew the test set at random with replacement 10,000 times, scored both methods on each redrawn set, and took the middle 95\% of the resulting differences as the 95\% CI. A CI that excludes 0 indicates a clear difference; one that includes 0 means the two methods cannot be told apart. On the external datasets we redrew whole food series, objects or trays rather than single photographs, because photographs of the same plate are not independent. Post hoc analyses were declared in writing before they were run.

\section{Results}

\subsection{Accurate in Domain, No Better Than a VLM Elsewhere}
\label{sec:zeroshot}

On the Nutrition5k validation set, the fine-tuned DINOv2 reached an MAE of 24.8\g, against 60.3\g for Qwen (difference $-35.6$\g; 95\% CI $-41.3$ to $-30.3$) and 56.5\g for \texttt{gpt-6-luna} (Table~\ref{tab:zeroshot}); 78.5\% of its estimates were within 20\% of the true mass, against 37.3\% for Qwen. These errors are close to those of published supervised networks on the same official test split (Appendix~\ref{sec:decision}).

\begin{figure*}[t]
\centering
\includegraphics[width=\textwidth]{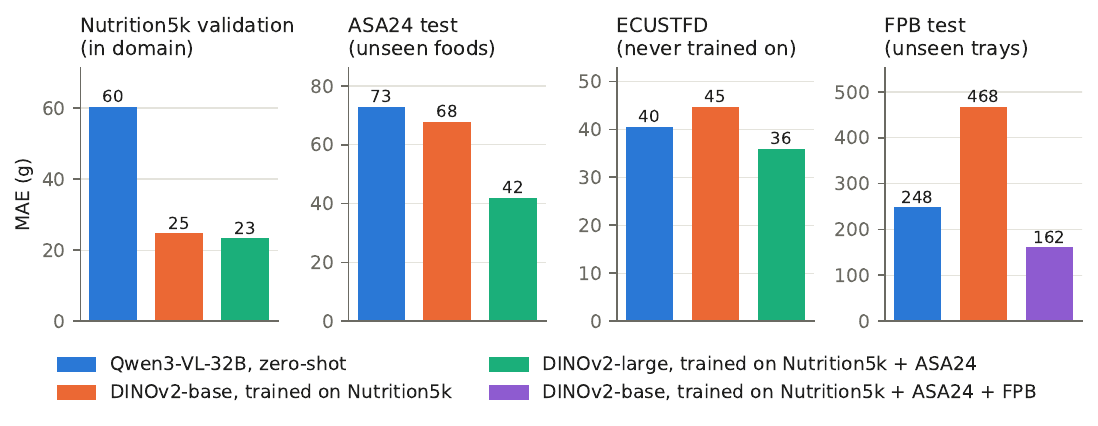}
\caption{MAE on the four datasets; note the different scales. Each color is one model, named by the datasets whose training photographs it used. The test photographs of ASA24 and FPB show foods and trays absent from their training photographs, and ECUSTFD was never used in training.}
\label{fig:crossdataset}
\end{figure*}

Without training photographs from the dataset being tested, this advantage disappeared. On ASA24 the fine-tuned model reached 67.9\g and Qwen 72.9\g ($-5.0$\g; $-13.0$ to $+2.9$), and on ECUSTFD 44.7 and 40.5\g ($+4.2$\g; $-2.3$ to $+10.7$). On FPB it failed: its MAE of 467.5\g was worse than always predicting the FPB training median, 373.3\g, whereas Qwen reached 248.5\g. \texttt{gpt-6-luna} behaved much like Qwen, with lower MAE on ECUSTFD and FPB and higher MAE on ASA24, so this comparison does not depend on the choice of VLM (Fig.~\ref{fig:crossdataset}).

Neither Qwen variant changed this picture. Given the ingredient names, Qwen was no more accurate: on the Nutrition5k validation set it reached 64.4\g with the true names and 70.5\g when choosing names from the dataset's ingredient list, against 60.3\g from the photograph alone, and on the external datasets the true food names changed its error by 11\g at most (Appendix~\ref{app:twostep}). A training-free correction, which rescales Qwen's estimate by its errors on the 20 most similar Nutrition5k training dishes, cut its validation error to 47.7\g but raised its ASA24 error from 72.9 to 76.2\g, because ASA24 photographs resemble no Nutrition5k dish (Appendix~\ref{sec:knn}).

\begin{table}[t]
\centering
\caption{MAE (g) on the Nutrition5k validation set and on the three external test sets without training on them. Constant, the median weight of the training set or, for ECUSTFD, of the test set itself.}
\label{tab:zeroshot}
\footnotesize
\setlength{\tabcolsep}{4pt}
\begin{tabular}{lrrrr}
\toprule
Method & N5k & ASA24 & ECUSTFD & FPB \\
\midrule
Qwen3-VL-32B & 60.3 & 72.9 & 40.5 & 248.5 \\
\texttt{gpt-6-luna} & 56.5 & 81.6 & 37.3 & 225.6 \\
Fine-tuned DINOv2-base & \textbf{24.8} & 67.9 & 44.7 & 467.5 \\
Constant prediction & 116.4 & 121.2 & 69.2 & 373.3 \\
\bottomrule
\end{tabular}
\end{table}

\subsection{The Gram Scale Fails in Two Ways}
\label{sec:relabs}

\begin{figure*}[t]
\centering
\includegraphics[width=\textwidth]{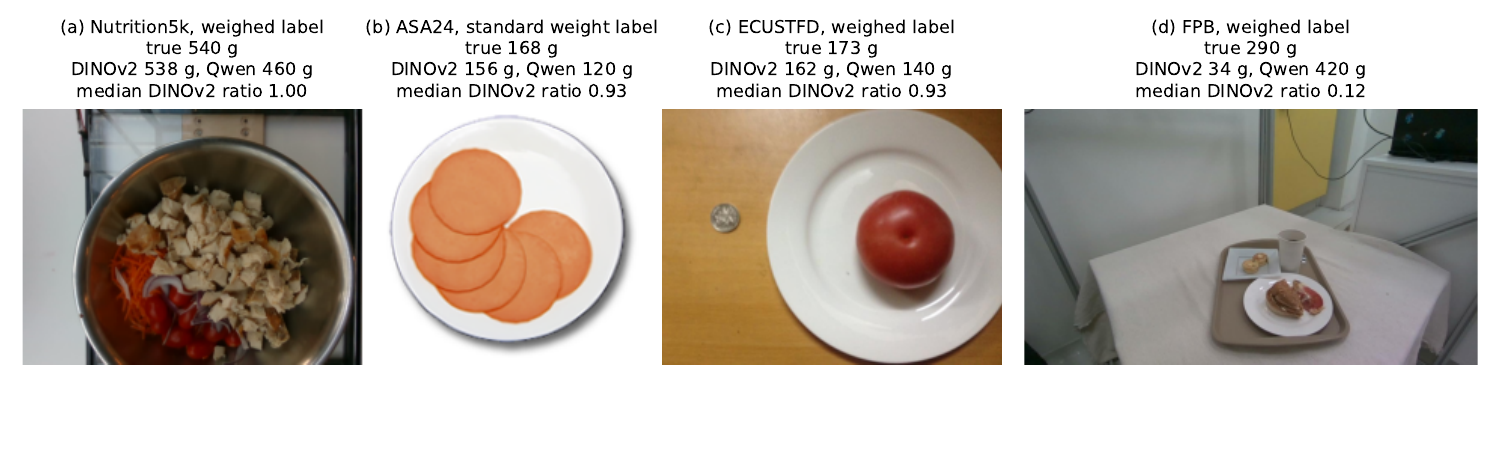}
\caption{A typical photograph from each dataset, with its true mass and the estimates of the Nutrition5k-only DINOv2 and Qwen. Each is the photograph whose ratio of DINOv2's estimate to the true mass is closest to the median ratio on that dataset, so it shows a typical error rather than a failure.}
\label{fig:datasets}
\end{figure*}

A model can rank photographs correctly, smallest to largest, and still give the wrong number of grams. Table~\ref{tab:relabs} separates the two, and Fig.~\ref{fig:datasets} shows a typical photograph from each dataset. In domain, the fine-tuned model did both well, with a Spearman correlation of 0.98 and a scale ratio of 1.00 on the Nutrition5k internal test set. Out of domain, it kept much of the ranking but lost the gram scale in two ways.

\emph{The scale shifted with framing.} A single RGB photograph carries no physical scale, so a model trained on one fixed overhead camera learns how many grams correspond to the visible food at that camera's distance and angle. On FPB trays, photographed from a distance and often at an angle, the Nutrition5k-only model predicted 0.14 times the true mass, and on ECUSTFD side views 0.76 times. The 25\,mm coin in every ECUSTFD photograph gives the width of the scene: for the same object, a wider scene gave a lower estimate from the fine-tuned models but not from Qwen. Rescaling the photographs to the scene width of Nutrition5k's camera did not fix the scale; it turned the error into another constant factor, which still had to be learned from labeled photographs (Appendix~\ref{sec:framing}). The scale ratio near 1 on ECUSTFD top views therefore reflects similar framing rather than a learned physical scale.

\begin{table}[t]
\centering
\caption{Scale ratio of the Nutrition5k-only model for single ASA24 test foods, over each food's series of portions.}
\label{tab:perfood}
\footnotesize
\begin{tabular}{lr}
\toprule
Food & Scale ratio \\
\midrule
Yogurt, fruit, 4-ounce container & 0.21 \\
White rice & 0.36 \\
Cream of tomato soup & 0.37 \\
Median of 162 foods & 0.88 \\
Pita bread & 2.67 \\
Saltine crackers & 3.80 \\
Apple Cinnamon Cheerios & 6.87 \\
\midrule
All 1,000 photographs & 0.96 \\
\bottomrule
\end{tabular}
\end{table}

\emph{The scale differed from food to food.} On ASA24, whose overhead studio photographs resemble Nutrition5k's, the Nutrition5k-only model was right on average, with a scale ratio of 0.96, but this average hid large errors in both directions (Table~\ref{tab:perfood}). Computed for each of the 162 test foods over its series of portions, the scale ratio was below 0.67 for 30\% of foods and above 1.5 for 25\%: the model predicted about a third of the true mass for white rice and nearly four times the true mass for saltine crackers. Light, bulky foods such as cereal and crackers tended to be overestimated, and dense foods such as yogurt and rice underestimated, consistent with a model that sees how much food is visible but not its density. Because the factor differed from food to food, no single correction could fix it: dividing every prediction by 0.96 left MAE at 68.4\g, against 67.9\g before. Within each food, the model ranked the portions well, with a Spearman correlation of 0.86, but a count of the pixels that are not white plate ranked them as well, 0.90 (difference $-0.04$; $-0.09$ to $+0.02$). The portions of one food on one plate differ mainly in visible area, so this ranking shows that the model responds to visible food, not that it knows portion sizes.

The gram scale itself comes from fine-tuning. A head trained on frozen DINOv2 features still ranked photographs by size, though less well, but predicted only 0.28 times the true mass on ASA24 and 0.38 times on ECUSTFD, in every one of four seeds; a fine-tuned ResNet-50 kept most of the scale (Appendix~\ref{sec:drivers}). Qwen's scale was more uniform across datasets, with ratios of 0.69 to 1.10, but its errors in domain were larger.

\begin{table*}[t]
\centering
\caption{Ordering versus gram scale. Rescaled MAE, MAE after dividing every prediction by the scale ratio.}
\label{tab:relabs}
\footnotesize
\begin{tabular}{llrrrr}
\toprule
Test set & Method & Spearman & Scale ratio & MAE (g) & Rescaled MAE (g) \\
\midrule
Nutrition5k (internal) & Qwen3-VL-32B, single prompt & 0.88 & 1.10 & 54.3 & 54.0 \\
 & DINOv2-base, Nutrition5k only & 0.98 & 1.00 & 19.7 & 19.7 \\
 & DINOv2-large, Nutrition5k+ASA24 & 0.98 & 1.02 & 17.5 & 17.7 \\
\addlinespace
ASA24 & Qwen3-VL-32B, single prompt & 0.76 & 0.72 & 72.9 & 69.3 \\
 & DINOv2-base, Nutrition5k only & 0.79 & 0.96 & 67.9 & 68.4 \\
 & DINOv2-large, Nutrition5k+ASA24 & 0.90 & 1.02 & 42.0 & 42.2 \\
\addlinespace
ECUSTFD, top view & Qwen3-VL-32B, single prompt & 0.81 & 0.92 & 40.5 & 38.8 \\
 & DINOv2-base, Nutrition5k only & 0.75 & 1.02 & 44.7 & 45.2 \\
 & DINOv2-large, Nutrition5k+ASA24 & 0.83 & 1.10 & 36.0 & 38.0 \\
\addlinespace
ECUSTFD, side view & Qwen3-VL-32B, single prompt & 0.80 & 1.03 & 40.2 & 40.8 \\
 & DINOv2-base, Nutrition5k only & 0.79 & 0.76 & 55.0 & 45.2 \\
 & DINOv2-large, Nutrition5k+ASA24 & 0.84 & 0.80 & 48.3 & 41.7 \\
\addlinespace
FPB & Qwen3-VL-32B, single prompt & 0.70 & 0.69 & 248.5 & 199.8 \\
 & DINOv2-base, Nutrition5k only & 0.56 & 0.14 & 467.5 & 431.5 \\
 & DINOv2-base, Nutrition5k+ASA24+FPB & 0.87 & 0.82 & 161.7 & 138.8 \\
\bottomrule
\end{tabular}
\end{table*}

\subsection{Only Same-Setting Data Restores the Scale}
\label{sec:multidomain}
\label{sec:fewshotresults}

\paragraph{Training photographs from the same dataset} Adding training photographs from the dataset being tested, of foods or trays absent from its test set, restored much of the accuracy (Table~\ref{tab:asa24}, Fig.~\ref{fig:crossdataset}). These results are same-dataset training, not cross-dataset transfer. On ASA24, DINOv2-large trained on Nutrition5k and the expanded ASA24 training set reduced test MAE from 67.9 to 42.0\g ($-25.9$\g; 95\% CI $-34.3$ to $-18.0$), although none of the 162 test foods appeared in training; all four two-domain models declared before the test set was scored reduced it by 22 to 26\g. On FPB, the three-domain model reduced MAE from 467.5 to 161.7\g, and its scale ratio rose from 0.14 to 0.82. The two-domain model also improved on ECUSTFD, which it never saw, from 44.7 to 36.0\g, and kept its accuracy on Nutrition5k, at 23.3\g.

Two controls show where these gains came from. A larger backbone alone did not help: DINOv2-large trained on Nutrition5k only had an MAE of 56.7\g on the 5,099 ASA24 training photographs, against 56.3\g for DINOv2-base, and 241.0 against 242.8\g on the FPB training photographs. Mixing datasets did not help either: DINOv2-base trained on FPB alone matched the three-domain model on FPB, 58.8 against 58.6\g in cross-validation ($+0.2$\g; $-1.1$ to $+1.4$). The gains therefore came from photographs of the same setting. Most of the remaining ASA24 error was specific to each food: removing each food's average log error, estimated from its other photographs, would reduce the two-domain model's ASA24 validation MAE from 37.2 to 21.8\g.

\begin{table}[t]
\centering
\caption{MAE (g) of fine-tuned DINOv2 by training data. \textsuperscript{T}, the model was trained on photographs from that dataset; *, not scored on the test set.}
\label{tab:asa24}
\footnotesize
\setlength{\tabcolsep}{4pt}
\begin{tabular}{llrrrr}
\toprule
Training data & Backbone & N5k & ASA24 & ECUSTFD & FPB \\
\midrule
N5k & Base & 24.8 & 67.9 & 44.7 & 467.5 \\
N5k + ASA24 & Large & \textbf{23.3} & \textbf{42.0}\textsuperscript{T} & \textbf{36.0} & 463.6 \\
N5k + ASA24 + FPB & Base & 24.6 & --* & -- & \textbf{161.7}\textsuperscript{T} \\
\bottomrule
\end{tabular}
\end{table}

\paragraph{One labeled photograph of a known food}\label{sec:fewshot} When the foods are known in advance, as on a cafeteria menu or a study's fixed food list, a few photographs of each food can be weighed once and used to correct the model. ASA24 allows this to be tested, because each food is photographed at several portions. For each test food we took $k=1$, 2 or 3 of its photographs as labeled, measured the model's error on them as the median log ratio of predicted to true weight, $e_f$, and multiplied the predictions for the food's other photographs by $\exp(-0.75\,e_f)$. If the model predicted 50\% too much on the labeled photograph, for example, the other predictions were lowered accordingly; the factor 0.75 applies only part of the correction because it rests on few photographs, and it was chosen on the ASA24 validation set. The model was the two-domain DINOv2-large, unchanged. The labeled photographs were drawn at random 20 times, and only the remaining photographs were scored.

This is known-food adaptation: the system must know the food in each photograph and hold labeled photographs of it. With one labeled photograph per food, calibration reduced ASA24 test MAE from 42.6 to 28.3\g (Table~\ref{tab:fewshot}); two and three photographs gave 27.3 and 26.7\g. The model added accuracy beyond the labeled photograph itself: scaling the labeled photograph's weight by the ratio of food areas, without the model, gave 48.0\g on the same draws, 19.6\g worse than the calibrated model (95\% CI 12.3 to 27.6).

\begin{figure*}[t]
\centering
\includegraphics[width=\textwidth]{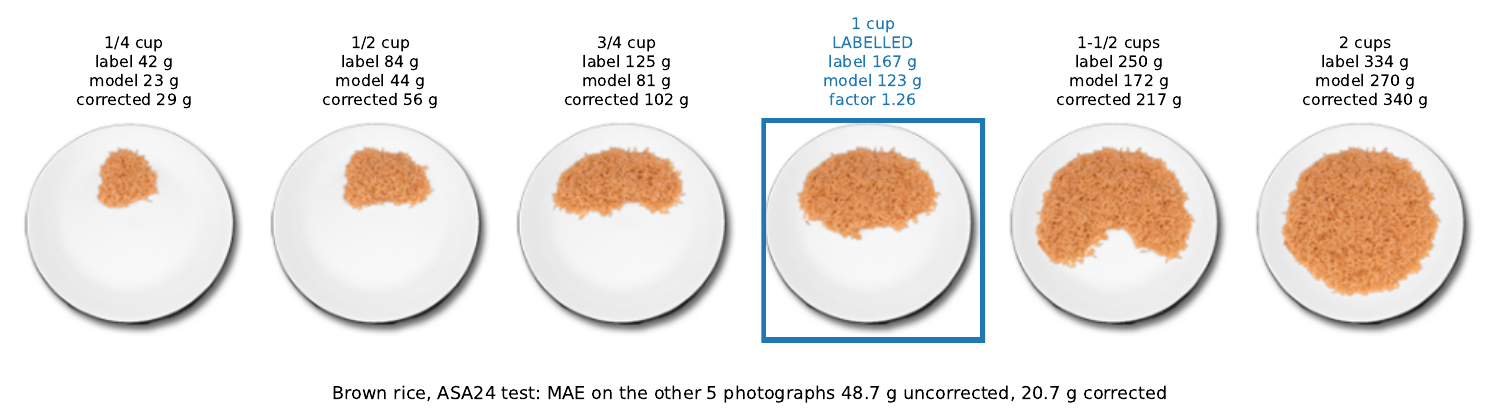}
\caption{Per-food calibration on one ASA24 test series (brown rice, aerial view; Nutrition5k+ASA24 DINOv2-large). The middle portion (blue) serves as the labeled photograph: the model predicted 123\g for a label of 167\g, giving a correction factor $\exp(-0.75\log(123/167))=1.26$, which is applied to the other five photographs. Labels are FNDDS standard-portion weights. Illustration: series with at least six photographs were ranked by how close their MAE reduction, with the middle portion as the labeled photograph, was to the median over series (56\%); this series (58\%) was the third closest and was chosen as a single, plainly visible food. In the evaluation (Table~\ref{tab:fewshot}) the labeled photograph is drawn at random.}
\label{fig:calexample}
\end{figure*}

Fig.~\ref{fig:calexample} shows why one photograph is enough on ASA24. The photographs of a food form one studio series: the same plate, the same overhead camera and, mostly, the same framing, with only the amount of food changing. The model ordered the brown-rice portions correctly but underestimated every one of them, predicting 0.53 to 0.81 times the labeled grams. That shortfall is largely a property of the food and its presentation, not of the portion, so the one labeled photograph measured it, and a single factor (1.26) corrected all the other portions, reducing their MAE from 48.7 to 20.7\g. This also marks the limit of the result: the labeled and corrected photographs shared plate, angle and lighting. A reference photograph taken in a different setting would also carry the framing shift described in Section~\ref{sec:relabs}, so the gain from per-food calibration in practice is likely to be smaller than on ASA24.

\begin{table}[t]
\centering
\caption{Per-food calibration on ASA24: MAE (g) with $k$ labeled photographs per food.}
\label{tab:fewshot}
\small
\begin{tabular}{lrrrr}
\toprule
& \multicolumn{2}{c}{Validation} & \multicolumn{2}{c}{Test} \\
$k$ & Uncal. & Cal. & Uncal. & Cal. \\
\midrule
1 & 37.7 & 28.7 & 42.6 & \textbf{28.3} \\
2 & 37.7 & 25.4 & 42.6 & 27.3 \\
3 & 37.7 & 25.2 & 43.0 & 26.7 \\
\bottomrule
\end{tabular}
\end{table}

\subsection{Supporting Findings}

\paragraph{Prediction intervals failed out of domain}\label{sec:intervals} A prediction interval is a range around an estimate that should contain the true mass with a stated probability. We gave every photograph a 90\% interval around its own estimate, from the estimate divided by a factor $k$ to the estimate times $k$. The factor was set once for each model, before testing, from its errors on held-out training data: $k$ is the 90th percentile of these errors, measured as a ratio of estimate to true mass. The construction resembles conformal prediction but does not carry its coverage guarantee~\cite{angelopoulos2023conformal} (Appendix~\ref{sec:uncertainty}). On a test set, the coverage is the share of photographs whose true mass fell inside their own interval; it should be close to 90\%. For the Nutrition5k-only model, $k$ was 1.35, so a 100\g estimate received the interval 74 to 135\g. This interval covered 89.8\% of the Nutrition5k validation dishes, as intended, but only 32.1\% of ASA24, 49.3\% of ECUSTFD and 0.9\% of FPB photographs (Table~\ref{tab:conformal}): on ASA24, the true mass fell outside the interval for two photographs in three, although the interval looked as confident as on Nutrition5k. On FPB, where estimates were about seven times too small, the interval was both narrow and in the wrong place. Calibrating the two-domain model on Nutrition5k and ASA24 errors restored coverage close to 90\% on ASA24 and ECUSTFD, but only by widening the interval to $k=2.07$, which nearly tripled its typical width on ASA24, from 58 to 156\g; on FPB, which was not part of the calibration, coverage stayed at 7.0\%. Intervals are therefore trustworthy only for photographs like those used to calibrate them. Qwen's self-reported confidence did not track its errors either.

\begin{table*}[t]
\centering
\caption{Coverage (\%) of 90\% prediction intervals, with 95\% CIs, and width, the median of upper minus lower bound in grams. An interval that works as intended has coverage near 90\%, with a CI that includes 90; observed coverage varies around 90 by chance. No validation value is given for the two-domain model, whose intervals were calibrated partly on the validation set.}
\label{tab:conformal}
\footnotesize
\setlength{\tabcolsep}{4pt}
\begin{tabular}{lrrrr}
\toprule
 & \multicolumn{2}{c}{DINOv2-base, Nutrition5k} & \multicolumn{2}{c}{DINOv2-large, Nutrition5k+ASA24} \\
\cmidrule(lr){2-3}\cmidrule(lr){4-5}
Test set & Coverage (95\% CI) & Width (g) & Coverage (95\% CI) & Width (g) \\
\midrule
Nutrition5k validation & 89.8 (87.1--92.4) & 101 & -- & -- \\
Nutrition5k test & 91.2 (88.6--93.6) & 95 & 99.2 (98.4--99.8) & 258 \\
ASA24 & 32.1 (26.3--37.9) & 58 & 88.7 (84.8--92.4) & 156 \\
ECUSTFD & 49.3 (42.2--57.0) & 71 & 91.3 (86.2--95.6) & 199 \\
FPB & 0.9 (0.4--1.5) & 28 & 7.0 (5.6--8.5) & 69 \\
\bottomrule
\end{tabular}
\end{table*}

\paragraph{A random Nutrition5k split overstated accuracy}\label{sec:siblings} Nutrition5k often photographed the same plate several times as ingredients were added, so two dishes captured seconds apart can show nearly the same food. We call two dishes siblings when they were captured within 120\,s and one's ingredient list contains the other's. The official split keeps siblings together, but our internal test set, drawn at random from the official training split, did not: 84\% of its dishes had a sibling in the training set, against 6\% of validation dishes. The fine-tuned DINOv2 was more accurate on test dishes with a training sibling, with 58\% against 40\% of its estimates within 10\% of the true mass, whereas Qwen, which is not trained on these dishes, showed no such difference. On the internal test set the fine-tuned model reached 19.7\g, against 24.8\g on validation, so we report the validation set as the in-domain result.

\section{Discussion}

\subsection{Implications}

Which estimator to use depends on what is available from the setting where it will be used (Appendix~\ref{sec:decision}). With training data from the same camera and menu, a fine-tuned image model was far more accurate than a zero-shot VLM. With no data from that setting it was no better, and on photographs framed very differently it was much worse. With labeled photographs from that setting, even of other foods or trays, the fine-tuned model was again the most accurate, and when the foods are known in advance, one weighed photograph of each food helped further. A model's accuracy in its own setting therefore says little about its accuracy elsewhere. A deployment needs labeled photographs from its own camera setting and, for a fixed list of foods, from its own foods, and its uncertainty estimates must be checked in that setting as well. The similarity of a photograph's frozen DINOv2 embedding to the training data flagged most external photographs as unfamiliar (Appendix~\ref{sec:uncertainty}); such a check can tell a system when new calibration data are needed, though not how wrong a given estimate is.

Zero-shot VLMs offer the opposite trade-off. Their errors were larger in domain but more uniform across datasets and did not depend on framing, which makes them a reasonable default when no labeled data from the setting of use exist. Supplying them with the ingredient names did not improve mass estimates (Appendix~\ref{app:twostep}).

The per-food result on ASA24 is likely an upper bound. Its labeled and corrected photographs share plate, angle and lighting; a reference photograph taken elsewhere would also carry the framing shift, so the gain in practice is likely to be smaller. More generally, a reference, whether an object of known size or a labeled photograph, reduced the scale error to a constant that still had to be learned in the setting of use.

\subsection{Limitations}

This study has five main limitations. First, the external datasets are curated: ASA24 labels are FNDDS standard portion weights rather than weighings, ASA24 photographs are $160\times160$-pixel studio thumbnails, ASA24 and ECUSTFD show single foods, and the fine-tuned models' fixed $448\times336$ input distorts the square ASA24 and portrait FPB photographs. Second, we compared the fine-tuned model with zero-shot VLMs that were not trained on weighed photographs; fine-tuning a VLM might narrow the gap, and larger proprietary VLMs might behave differently. Third, the official Nutrition5k test split served as our validation set and was used for model selection, so no result here is a clean comparison with published official-split results, and our internal test set shares near-duplicate plates with the training set. Fourth, every test-set result comes from one training seed; seed variability was measured for three Nutrition5k-only variants on data that are not test sets. Fifth, adding a dataset to training changed training-set size, data composition and optimizer steps together; the controls rule out backbone size and dataset mixing as explanations, but learning curves in the amount of same-dataset training data are missing. These are retrospective results on public images and do not show whether a dietary assessment system built on them improves intake estimates.

\section{Conclusion}

A single-image food mass estimator fine-tuned on weighed food was accurate in its own setting but did not carry its gram scale to new datasets. Without training data from the setting of use it was no more accurate than a zero-shot VLM; its scale shifted with framing and differed from food to food, and the ordering it kept was no better than a count of food pixels. Labeled photographs from the same setting, of other foods or trays, restored much of the accuracy, and one labeled photograph per known food restored more, whereas a larger backbone did not. Prediction intervals calibrated in domain failed outside it, and random Nutrition5k splits overstate accuracy. Single-image mass estimators should therefore be validated and calibrated in the setting where they will be used.

\section*{Data Availability}
All datasets are publicly available from their original sources~\cite{thames2021nutrition5k,nci_asa24_images,usda_fndds,liang2017ecustfd,fpb2025}. The lists of dishes and photographs in each of our sets, the analysis code, the dated analysis plans and the trained models are released at \url{https://github.com/yongshenglian/foodmass-transfer}.

\bibliographystyle{IEEEtran}
\bibliography{references}

\appendices
\section{Datasets and Labels}
\subsection{ASA24 Labels and Audit}
\label{sec:labelaudit}

\begin{figure*}[t]
\centering
\includegraphics[width=0.78\textwidth]{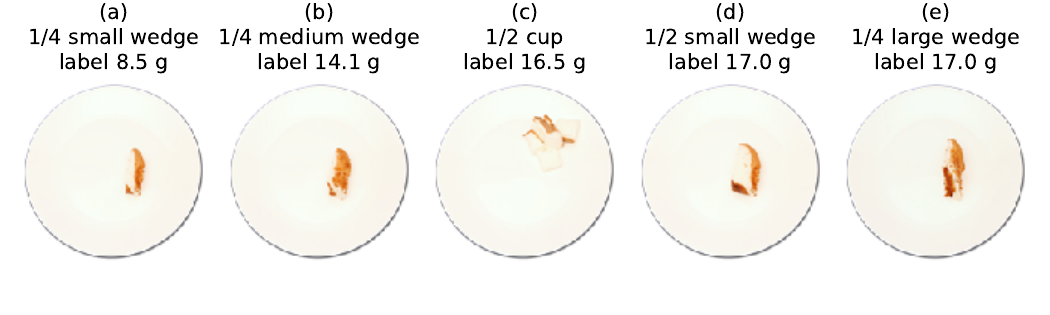}
\caption{One ASA24 test series, angel food cake, shown at its native $160\times160$ pixels. Each label is the FNDDS weight of the stated portion, not a weighing, so (d) and (e) receive the same label although they show different amounts.}
\label{fig:asa24series}
\end{figure*}

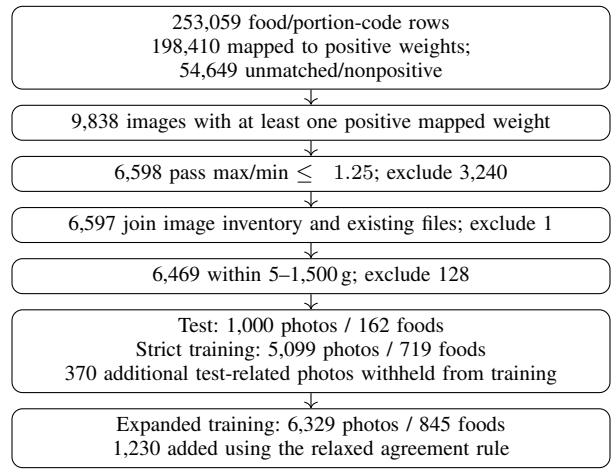
\begin{figure}[t]
\centering
\begin{tikzpicture}[node distance=2mm, every node/.style={draw, rounded corners, align=center, font=\scriptsize, text width=7.7cm, inner sep=3pt}]
\node (a) {253,059 food/portion-code rows\\198,410 mapped to positive weights; 54,649 unmatched/nonpositive};
\node[below=of a] (b) {9,838 images with at least one positive mapped weight};
\node[below=of b] (c) {6,598 pass max/min $\leq1.25$; exclude 3,240};
\node[below=of c] (d) {6,597 join image inventory and existing files; exclude 1};
\node[below=of d] (e) {6,469 within 5--1,500\,g; exclude 128};
\node[below=of e] (f) {Test: 1,000 photos / 162 foods\\Strict training: 5,099 photos / 719 foods\\370 additional test-related photos withheld from training};
\node[below=of f] (g) {Expanded training: 6,329 photos / 845 foods\\1,230 added using the relaxed agreement rule};
\draw[->] (a)--(b);\draw[->] (b)--(c);\draw[->] (c)--(d);\draw[->] (d)--(e);\draw[->] (e)--(f);\draw[->] (f)--(g);
\end{tikzpicture}
\caption{ASA24 label construction. Agreement concerns successfully mapped positive weights; unmapped rows do not automatically exclude an image. Expanded training retains the test-food exclusions. Labels are standard-portion assignments, not physical measurements.}
\label{fig:asa24flow}
\end{figure}

\paragraph{ASA24 labels} ASA24's metadata use FNDDS 2007--2008 food codes, whereas we took portion weights and energy values from FNDDS 2011--2012; 78\% of the photographs' food-and-portion code pairs were found in that release, and photographs whose codes were not found were left unlabeled (Fig.~\ref{fig:asa24flow}). A photograph linked to several foods was labeled with the median of their weights when the heaviest was no more than 25\% above the lightest; ``Chocolate drink powder, 1/4 cup'', linked to cocoa, milk and malted milk powders weighing 44.0, 36.4 and 19.2\g, was therefore excluded. The expanded training set instead kept a photograph when at least 60\% of its linked foods gave weights within 25\% of their median, which added 1,230 training photographs; test photographs were always labeled with the strict rule. Because each label is the FNDDS weight of a stated portion, different amounts of one food can share a label (Fig.~\ref{fig:asa24series}).

\paragraph{Sensitivity} The comparisons did not depend on the labeling rule. Restricting the test set to the 549 photographs of 97 foods linked to a single food code gave MAEs of 65.3\g for the Nutrition5k-only DINOv2, 38.1\g for the two-domain DINOv2 and 64.0\g for Qwen, against 67.9, 42.0 and 72.9\g overall. Tightening the agreement rule from 25\% to 0\% kept 805 test photographs, with MAEs of 66.0, 43.8 and 68.3\g. In training, the expanded rule reduced test MAE by 0.8\g for DINOv2-base and 1.0\g for DINOv2-large, with 95\% CIs spanning zero.

\paragraph{Audit} Among the ten largest two-domain ASA24 errors, the breaded-swordfish series shows fewer pieces as the labeled amount increases, which suggests an image--metadata mismatch that we cannot resolve; we kept its labels. For FPB, all 11,541 retained photographs reproduced their summed masses from the original item labels. Excluding the 181 evaluated photographs with a nonpositive item weight changed FPB MAE by at most 3.4\g for any model.

\subsection{Nutrition5k Internal Test Set}
\label{app:sampling}

The internal test set was drawn on 12 September 2026 from the 2,758 dishes of Nutrition5k's official training split. Three dishes without an overhead photograph were excluded, and so were 679 dishes already used in exploratory experiments; these remained available for training. Of the 2,076 eligible dishes, 502 were drawn by proportional stratified sampling over 20 strata of dish mass and ingredient count, and 498 of them were scored; the four set aside were chosen by a hash of a fixed seed and the dish ID, which reads no label, image or result. The remaining 2,253 dishes form the training set.

The internal test set was scored six times, each time only for models whose settings were already fixed. No model was trained, tuned or selected on test dishes, but the later evaluations were made with knowledge of the earlier results; the photograph-only models, for example, were specified after the first two evaluations had been seen. We therefore treat the internal test results as supporting rather than confirmatory.

\section{Fine-Tuned DINOv2: Training, Controls and Validation Results}
\subsection{Training}
\label{app:training}

Table~\ref{tab:training} lists every setting. Photographs were resized to $448\times336$ pixels, a 4:3 shape whose sides are multiples of DINOv2's 14-pixel patches. ASA24 cut-outs were first placed on white, and FPB photographs were first reduced to at most 1,024 pixels. Photographs were never cropped or zoomed, because the apparent size of the food tells the model about its amount. The main target was log mass, so a 20\% miss counts the same for a 50\g and a 500\g dish. The depth target, available for Nutrition5k only, was a height-sum proxy computed from the overhead depth map. The table height was the median depth of a 20-pixel border. Food heights above it were clipped to 0--15\,cm, a 2\,cm plate layer was subtracted, and the heights were summed, log-transformed and standardized. This proxy correlates with log mass at 0.33 and is not a physical volume.

The five copies of each model were trained on four of five folds each. Nutrition5k folds were random. ASA24 folds were grouped by food, and FPB folds by annotation signature, so each copy was evaluated on held-out foods or trays; these out-of-fold errors guided every modeling choice. A new photograph's estimate is the mean of the five copies' log predictions, converted to grams and limited to 10--1,500\g. All final models used seed 20261004, and the seed controls used 20261011--20261013. Training used PyTorch 2.6 and Hugging Face Transformers 5.18 on one NVIDIA RTX A6000. A five-copy DINOv2-base run on Nutrition5k took about 27 minutes, and a five-copy DINOv2-large run on Nutrition5k and ASA24 about 3.7 hours. Qwen was served with vLLM 0.30 and never fine-tuned.

\begin{table}[!t]
\centering
\caption{Training settings for the DINOv2 models.}
\label{tab:training}
\footnotesize
\begin{tabular}{L{4.5cm}L{10cm}}
\toprule
Setting & Value \\
\midrule
Starting weights & DINOv2-base (86\,M parameters) or DINOv2-large (300\,M), as released by Meta AI \\
Input & RGB photograph, $448\times336$ pixels, ImageNet normalization \\
Augmentation (training only) & Horizontal and vertical flips ($p=0.5$ each); brightness $\times U(0.9,1.1)$ and $+U(-0.05,0.05)$; no cropping \\
Image features & [CLS] token concatenated with mean patch token \\
Prediction layers & 512-unit GELU layer, dropout 0.2; linear outputs for log weight (and nutrients); linear depth output from the image features \\
Main loss & Smooth-$\ell_1$ on log weight, $\beta=0.1$ \\
Depth loss (training only) & Smooth-$\ell_1$ on the standardized height-sum proxy, $\beta=1$, weight 1.0; Nutrition5k photographs only \\
Nutrient loss (optional) & Smooth-$\ell_1$ on $\log(1+y)$ for energy, fat, carbohydrate, protein, $\beta=0.1$, weight 1.0 \\
Optimizer & AdamW, weight decay 0.05 \\
Learning rates (peak) & DINOv2 layers $2\times10^{-5}$; new layers $10^{-3}$ \\
Schedule & One-cycle, 10\% warm-up, then decay \\
Batch size; passes & 16 photographs; 15 passes (epochs) \\
Gradient clipping; precision & Norm 1.0; bfloat16 \\
Cross-validation & 5 copies; Nutrition5k random, ASA24 grouped by food, FPB grouped by annotation signature \\
Prediction & Mean of the 5 copies' log weights; limited to 10--1,500\,g; photograph only \\
Hardware & One NVIDIA RTX A6000 (48\,GB) \\
\bottomrule
\end{tabular}
\end{table}

\subsection{Supervised Baselines and Controls}
\label{sec:drivers}
\label{sec:controls}

\paragraph{Matched baselines} Four variants were trained with the same photographs, head, augmentation, schedule, seed and five-fold averaging as the Nutrition5k-only model: DINOv2-base fine-tuned without the depth target; DINOv2-base frozen, with only the head trained, with and without the depth target; and ResNet-50 pretrained on ImageNet~\cite{he2016resnet}, fine-tuned with the depth target. ResNet-50 was trained at four backbone learning rates, and the best on validation is reported. Table~\ref{tab:matched} gives the results.

Fine-tuning the whole backbone mattered most. With the backbone frozen, the model was 10.4\g worse on the validation set (95\% CI 7.6 to 13.2). Out of domain it still ordered photographs by size, but it lost the gram scale: it predicted 0.28 times the true mass on ASA24 and 0.38 times on ECUSTFD, against 0.96 and 1.02 for the fine-tuned model. A fine-tuned ResNet-50 was also 10.4\g worse in domain but largely kept the gram scale. The gram scale therefore comes from fine-tuning the backbone, and the accuracy from DINOv2's pretraining. The depth target gave a small in-domain gain, 0.6 to 0.9\g in every seed, and no consistent gain out of domain.

\begin{table*}[t]
\centering
\caption{Matched supervised baselines. $\Delta$, difference in MAE from the final model, with 95\% CI; $\rho_{\mathrm{food}}$, Spearman correlation within each ASA24 food; Scale, scale ratio; LR, backbone learning rate.}
\label{tab:matched}
\footnotesize
\setlength{\tabcolsep}{3.5pt}
\resizebox{\textwidth}{!}{%
\begin{tabular}{@{}lrlrlrrrlrrr@{}}
\toprule
& \multicolumn{2}{c}{Validation} & \multicolumn{4}{c}{ASA24} & \multicolumn{3}{c}{ECUSTFD} & \multicolumn{2}{c@{}}{FPB} \\
\cmidrule(lr){2-3}\cmidrule(lr){4-7}\cmidrule(lr){8-10}\cmidrule(l){11-12}
Model & MAE & $\Delta$ (95\% CI) & MAE & $\Delta$ (95\% CI) & $\rho_{\mathrm{food}}$ & Scale & MAE & $\Delta$ (95\% CI) & Scale & MAE & Scale \\
\midrule
DINOv2, fine-tuned, depth (final) & 24.8 & --- & 67.9 & --- & 0.86 & 0.96 & 44.7 & --- & 1.02 & 468 & 0.14 \\
DINOv2, fine-tuned, no depth & 25.7 & $+0.9$ (0.2, 1.6) & 70.4 & $+2.5$ (0.0, 5.2) & 0.86 & 1.00 & 44.4 & $-0.3$ ($-1.7$, 1.1) & 0.99 & 468 & 0.14 \\
DINOv2, frozen, depth & 35.2 & $+10.4$ (7.6, 13.2) & 115.7 & $+47.8$ (36.7, 58.5) & 0.82 & 0.28 & 88.2 & $+43.5$ (32.7, 54.1) & 0.38 & 501 & 0.07 \\
DINOv2, frozen, no depth & 35.5 & $+10.8$ (7.9, 13.6) & 116.4 & $+48.5$ (37.3, 59.2) & 0.79 & 0.28 & 86.1 & $+41.4$ (30.8, 51.8) & 0.41 & 501 & 0.07 \\
ResNet-50, depth, LR $10^{-3}$ & 35.2 & $+10.4$ (8.0, 13.1) & 78.3 & $+10.4$ (3.4, 17.1) & 0.80 & 0.82 & 60.0 & $+15.3$ (10.6, 20.0) & 1.14 & 471 & 0.14 \\
\bottomrule
\end{tabular}}
\end{table*}

\paragraph{Seed control} Each model above was trained with one seed. The final model, its version without depth and its frozen version were retrained with three more seeds and evaluated on photographs that were neither training nor test data for them: the Nutrition5k validation set and the ASA24 and FPB training photographs. The frozen model lost the scale in all four seeds (Table~\ref{tab:seeds}).

\begin{table}[t]
\caption{Seed controls: mean (range) over four seeds. FT, fine-tuned; $\rho$, Spearman correlation.}
\label{tab:seeds}
\centering
\footnotesize
\setlength{\tabcolsep}{1pt}
\begin{tabular}{@{}lccc@{}}
\toprule
 & FT, depth & FT, no depth & Frozen, depth \\
\midrule
\multicolumn{4}{@{}l}{\emph{Nutrition5k validation (502 dishes)}} \\
MAE (g) & 24.8 (24.6--24.9) & 25.6 (25.3--25.8) & 35.0 (34.3--35.4) \\
\addlinespace
\multicolumn{4}{@{}l}{\emph{ASA24 training photographs (5,099)}} \\
MAE (g) & 56.3 (55.8--56.8) & 57.1 (56.7--57.9) & 99.9 (98.5--100.5) \\
Scale & 0.99 (0.94--1.04) & 1.03 (0.96--1.11) & 0.30 (0.30--0.32) \\
$\rho$ & 0.80 (0.80--0.81) & 0.80 (0.79--0.80) & 0.79 (0.79--0.80) \\
Within-food $\rho$ & 0.87 (0.86--0.87) & 0.86 (0.85--0.87) & 0.76 (0.75--0.77) \\
\addlinespace
\multicolumn{4}{@{}l}{\emph{FPB training photographs (9,190)}} \\
MAE (g) & 242.8 (241.2--244.7) & 241.4 (240.7--242.5) & 267.2 (266.8--268.0) \\
Scale & 0.21 (0.19--0.22) & 0.22 (0.21--0.22) & 0.10 (0.10--0.10) \\
$\rho$ & 0.48 (0.47--0.49) & 0.51 (0.49--0.52) & 0.48 (0.47--0.49) \\
\bottomrule
\end{tabular}
\end{table}

\paragraph{Two-domain models and backbone} Four two-domain models were declared before the ASA24 test set was scored, and each was scored once: DINOv2-base (45.8\g), DINOv2-large (43.0\g), DINOv2-large with the expanded training set (42.0\g) and the ensemble of the two DINOv2-large models (42.4\g). All four reduced ASA24 MAE by 22--26\g relative to the Nutrition5k-only model (67.9\g). We report the DINOv2-large with the expanded set, which had the lowest validation MAE of the single models (37.2\g). The larger backbone alone did not improve transfer: DINOv2-large trained on Nutrition5k only reached 56.7\g on the ASA24 training photographs, against 56.3\g for DINOv2-base. The gain on ASA24 therefore came from the ASA24 training photographs. On Nutrition5k, the larger backbone reduced validation error by about 1.8\g, and the ASA24 data neither helped nor hurt.

\paragraph{FPB-only control} The three-domain model reached 161.7\g on FPB, against 467.5\g for the Nutrition5k-only model. The gain came from the FPB photographs alone: DINOv2-base trained only on them reached an out-of-fold MAE of 58.8\g on the FPB training photographs, against 58.6\g for the three-domain model ($+0.2$\g; 95\% CI $-1.1$ to $+1.4$).

\subsection{Nutrition5k Validation Results}
\label{sec:n5kresults}

The training, validation and internal test sets have 2,253, 502 and 498 dishes, with mean masses of 223, 200 and 195\g; Appendix~\ref{app:sampling} describes how the internal test set was drawn. We report the validation set as the in-domain result because the internal test set shares near-duplicate plates with training (Section~\ref{sec:siblings}). Its errors are slightly optimistic, because models were chosen on it. Table~\ref{tab:testphoto} gives the results.

\begin{table*}[t]
\centering
\caption{Mass error on the Nutrition5k validation set. $\Delta$, difference in MAE from single-prompt Qwen, with 95\% CI.}
\label{tab:testphoto}
\resizebox{\textwidth}{!}{%
\begin{tabular}{lrrrrrr}
\toprule
Method & MAE (g) & Median AE (g) & RMSE (g) & Bias (g) & Within 50\,g & $\Delta$ MAE vs single-prompt Qwen (95\% CI), g \\
\midrule
Qwen3-VL-32B, single prompt & 60.35 & 41.00 & 94.34 & $-9.59$ & 60.4\% & -- \\
Qwen3-VL-32B, single prompt + kNN residual correction & 47.69 & 28.66 & 73.57 & $-0.50$ & 66.5\% & $-12.66$ ($-17.32$, $-8.41$) \\
Fine-tuned DINOv2-base, Nutrition5k only & 24.78 & 16.02 & 39.35 & $-4.33$ & 88.6\% & $-35.57$ ($-41.27$, $-30.26$) \\
Fine-tuned DINOv2-large, Nutrition5k+ASA24 & 23.29 & 14.08 & 38.49 & $-0.48$ & 90.2\% & $-37.06$ ($-42.90$, $-31.61$) \\
\bottomrule
\end{tabular}}
\end{table*}

\section{Qwen: Prompts, Variants and kNN Correction}
\subsection{Prompts}
\label{app:prompts}

\paragraph{System prompt} Sent with every request.
\begin{lstlisting}
You are an expert at analyzing images with computer vision. I will present you with a picture of food, which might be placed on a plate, inside a spoon, or contained within different vessels. Your job is to accurately identify the food depicted in the image.
\end{lstlisting}

\paragraph{User prompt} Used for every photograph-only result.
\begin{lstlisting}
The image displays one serving of food. Identify each food item in the image yourself. Make a best-effort integer gram estimate for EACH food item, including ingredients that are mixed in, hidden, or present in small amounts. Also report one dish-level confidence score from 0 to 100. Define confidence as your estimated percent probability that the SUM of your ingredient estimates is within 50 grams of the true total food weight. Judge confidence from visible boundaries, portion separability, available scale cues, and uncertainty from hidden ingredients. Do not inflate confidence. Use 90-100 only for unusually clear portions with reliable scale cues; 80-89 for mostly clear portions with moderate ambiguity; 60-79 for mixed foods, weak scale cues, or meaningful hidden ingredients; and 0-59 for severe ambiguity. Always provide the gram estimates even when confidence is low; filtering is performed externally. Respond with exactly these lines and no other text:
CONFIDENCE: integer%
food item name: integer g
(one line for every food item)
\end{lstlisting}

\paragraph{Retry instruction} Appended once after an invalid response.
\begin{lstlisting}
Your previous response was not in the required format. Include the confidence line and every listed ingredient exactly once.
\end{lstlisting}

\paragraph{Variants} When names were given, the first two sentences read ``The image displays one serving of a dish containing these known ground-truth ingredients: [INGREDIENT NAMES].'' and the model estimated grams for each listed ingredient. The two-step mode first asked: ``List every ingredient in this dish, including ingredients that are mixed in, hidden, or present in small amounts. Choose names only from this list, copied exactly: [555 NUTRITION5K INGREDIENT NAMES]. Respond with one ingredient name per line and no other text.''

\paragraph{Prompt wording on FPB} Qwen's underestimation of trays was not caused by the words ``one serving''. On 1,200 FPB training photographs, half of them with several items, replacing that opening with ``The image displays food; it may be a single item or a whole meal with several items'' changed the scale ratio on photographs with several items from 0.71 to 0.74 and reduced their MAE from 274.9 to 254.5\g ($-20.3$\g; 95\% CI $-27.4$ to $-13.4$); on single items MAE did not change ($-3.8$\g; $-12.9$ to $+5.1$).

\subsection{Qwen Given an Ingredient List or Names}
\label{app:twostep}

In the \emph{two-step} mode, Qwen first listed the ingredients it saw, choosing only from the 555 Nutrition5k ingredient names, and then estimated grams for each recognized name with the names prompt. The list is the same for every dish, but it is dataset-level vocabulary that a deployed system would not have, so this mode was run on Nutrition5k only. On the Nutrition5k validation set it reached 70.45\g, 10.11\g worse than the single prompt (95\% CI 5.85 to 14.43). It recalled 23\% of the annotated ingredient names, missing mostly oils, seasonings and hidden components.

Given the dish's true ingredient names, Qwen reached 64.37\g on the validation set, 4.02\g worse than the single prompt ($-0.17$ to $+8.35$), and 57.8 against 54.3\g on the internal test set. Given the true food name on ASA24 or the item names on ECUSTFD and FPB, it reached 69.7, 40.8 and 237.1\g, against 72.9, 40.5 and 248.5\g from the photograph alone. Names therefore did not improve mass estimates. They did matter for calories computed from Qwen's ingredient list: on the validation set, this calorie error rose from 81\,kcal with the true names to 100\,kcal with the names Qwen recognized itself (Appendix~\ref{sec:nutrients}). Evaluations that supply the true names~\cite{rodriguez2025chatgpt5,vlm2026comparative} thus overstate the accuracy of nutrients computed from the VLM's ingredient list, although not, in our data, the accuracy of mass.

Nutrition5k runs with the true names used bf16 weights; all other runs used the official FP8 checkpoint. Run with both on the validation set with names given, the two reached 64.37 and 63.67\g ($+0.70$\g; $-0.92$ to $+2.33$).

\subsection{kNN Residual Correction}
\label{sec:knn}
\label{app:knn}

This training-free baseline corrects Qwen by its errors on similar-looking dishes. Qwen was run on the 2,253 training dishes with their annotated names, so its log error $r_i=\log(m_i/q_i)$ on each training dish $i$ is known, where $m_i$ is the measured and $q_i$ the estimated mass. For a new photograph $x$, a frozen, off-the-shelf DINOv2-base embedding finds the 20 training dishes with the highest cosine similarity $s_i(x)$. Their errors are combined by a weighted median $\delta(x)$, with weights $w_i(x)=\exp\{[s_i(x)-\max_j s_j(x)]/\tau(x)\}$, where $\tau(x)$ is the larger of 0.05 and the standard deviation of the 20 similarities. The corrected estimate is
\begin{equation}
\hat{m}_{\mathrm{kNN}}(x) = \operatorname{clip}\!\left[\, q(x)\, e^{0.75\,\delta(x)},\ 10,\ 1500 \right].
\end{equation}
A median is used because Qwen's errors are heavy-tailed: it missed by more than a factor of 3 on 2.3\% of training dishes. The neighbor count and the factor 0.75 were fixed on the validation set. The approach follows analog-ensemble post-processing in weather forecasting~\cite{delle2013analog} and is related to localized conformal prediction~\cite{guan2023localized}.

On the Nutrition5k validation set, the correction reduced Qwen's MAE from 60.35 to 47.69\g ($-12.66$\g; 95\% CI $-17.32$ to $-8.41$). It works mainly because Qwen misjudges similar dishes in the same way, not because similar dishes weigh the same. With the true names, predicting each dish's mass as the weighted median of its neighbors' measured masses gave 56.72\g, whereas correcting Qwen's estimate of 63.67\g gave 48.87\g. Out of domain, the correction made Qwen worse: on ASA24 it raised the error from 72.9 to 76.2\g, because the median similarity to the nearest training dish was 0.49 there, against 0.81 within Nutrition5k.

\section{Further Analyses}
\subsection{Ordering and Framing}
\label{sec:framing}

The within-food ordering on ASA24 is not evidence of transferred portion knowledge, because a pixel count does as well. Counting the non-white pixels of each ASA24 cut-out gave a within-food Spearman correlation of 0.90 on the test set (95\% CI 0.85 to 0.94), against 0.86 for the Nutrition5k-only model and 0.72 for Qwen. Only the two-domain model, trained on ASA24 photographs of other foods, exceeded the pixel count, narrowly, at 0.94. Ordering the portions of one food on one plate is thus an easy task that visible food area largely solves.

The coin in ECUSTFD shows that the fine-tuned models' gram scale depends on framing. Within objects, where the true mass is constant, a wider scene lowered the estimate of the Nutrition5k-only model (slope of log error on log scene width $-0.33$; 95\% CI $-0.70$ to $-0.19$, top views) and of the two-domain model ($-0.40$; $-0.84$ to $-0.24$), but not of Qwen ($0.00$; $-0.09$ to $+0.09$). ECUSTFD scenes are about half as wide as Nutrition5k frames, with a median of 309 against 563\,mm. Rescaling every photograph to the scene width of a Nutrition5k frame therefore shrank the food, and the Nutrition5k-only model then predicted 0.47 times the true mass on top views and 0.33 times on side views. The scale ratio near 1 on unmatched top views thus reflects how ECUSTFD photographs happen to be framed. After matching, dividing by one factor estimated on other objects reduced MAE from 38.3 to 28.9\g for the two-domain model on top views ($-9.4$\g; $-13.8$ to $-5.5$). A reference object thus reduced the scale error to one constant, which still had to be learned from labeled photographs of the same setting. This follow-up was not planned, and the padding that matching requires was never seen in training, so these numbers are exploratory.

Camera angle mattered for the fine-tuned models but not for Qwen. On side views of the same ECUSTFD objects, the two-domain model's MAE rose from 36.0 to 48.3\g (paired difference per object $+8.0$\g; 95\% CI $+2.4$ to $+13.6$), whereas Qwen's was unchanged (40.5 and 40.2\g).

\subsection{Nutrients}
\label{sec:nutrients}

A mass estimate alone does not give calories; the composition of the dish is also needed. The nutrient head of the fine-tuned DINOv2 predicts energy, fat, carbohydrate and protein directly from the photograph. On the Nutrition5k validation set it reached an energy MAE of 41.4\,kcal, against a mean of 258\,kcal per dish, and 3.3, 4.4 and 3.8\g for fat, carbohydrate and protein. On ASA24 its energy MAE rose to 144.8\,kcal, so it did not transfer. The alternative converts total mass to nutrients through Qwen's ingredient list, which gives each ingredient's share of the dish. With the ingredients Qwen recognized itself, this gave 79--100\,kcal on validation even with the true total mass, because Qwen recognized only 23\% of the ingredient names.

Text-based food knowledge did not fix the scale either. We gave the two-domain DINOv2 three extra inputs on ASA24: Qwen's gram estimate, an encoding of the food name Qwen wrote, and USDA standard portion weights for that food, matched to FNDDS 2021--2023 with a sentence encoder~\cite{xiao2023bge}. None improved ASA24 validation MAE by more than 0.2\g, so none was scored on the test set.

\subsection{Uncertainty}
\label{sec:uncertainty}

\paragraph{Intervals} The interval for a new photograph was $[\hat m/k,\ \hat m k]$, where $\log k$ is the $\lceil (n+1)\alpha \rceil$-th smallest of $n$ calibration errors $|\log(\hat m/m)|$ at level $\alpha=0.9$. For the Nutrition5k-only model, the $n=2{,}253$ calibration errors were the out-of-fold errors on the training dishes. For the two-domain model, they pooled the ensemble's errors on the 502 validation dishes with the out-of-fold errors on the 5,099 ASA24 training photographs, $n=5{,}601$. For the three-domain model, they were the out-of-fold errors on the 9,190 FPB training photographs. These are not split-conformal intervals and carry no coverage guarantee~\cite{angelopoulos2023conformal}: most errors come from fold models rather than the deployed ensemble, and photographs of the same food or tray are not exchangeable. The 95\% CIs of coverage come from a bootstrap over dishes, food series, objects or trays. Calibrating the Nutrition5k-only model instead on the ensemble's validation errors gave nearly the same coverage: 91.6\%, 33.1\%, 50.7\% and 0.9\% on the internal test set, ASA24, ECUSTFD and FPB. Calibrated on FPB's own training photographs, the three-domain model's interval covered 78.1\% of FPB test photographs.

\paragraph{Qwen's confidence} Asked for the probability that its total was within 50\g, Qwen stated 74.5\%, 72.5\%, 88.0\% and 73.6\% on average on Nutrition5k, ASA24, ECUSTFD and FPB, whereas 60.8\%, 57.8\%, 67.0\% and 19.4\% of its answers were within 50\g. Its confidence barely separated answers within 50\g from the rest, with AUROC 0.43 to 0.57.

\paragraph{Unfamiliar photographs} The similarity of a photograph's frozen DINOv2 embedding to the nearest Nutrition5k training dish separated each external set from the internal test set, with AUROC 0.997 for ASA24, 0.978 for ECUSTFD and 0.997 for FPB. A threshold at the 5th percentile of the validation set flagged 1.4\% of internal test photographs but 95.5\% of ASA24, 74.2\% of ECUSTFD and 97.0\% of FPB photographs. Within ASA24, however, the similarity did not predict which photographs had large errors (Spearman correlation $-0.08$). Such a signal can tell a system that it is outside its validated domain, not how wrong an estimate is.

\section{Choosing an Estimator and Published Results}
\label{sec:decision}
\label{sec:context}

Table~\ref{tab:decision} summarizes which evaluated estimator was most accurate, given what is available from the setting of use.

\begin{table*}[t]
\centering
\caption{Most accurate estimator by the data available from the setting of use: MAE (g).}
\label{tab:decision}
\footnotesize
\begin{tabular}{L{3.6cm}L{3.2cm}L{5.6cm}}
\toprule
Available from the setting of use & Most accurate & Evidence \\
\midrule
Nothing & A zero-shot VLM, or no clear difference & ASA24: Qwen 72.9, fine-tuned 67.9 (difference not significant). ECUSTFD: 40.5 and 44.7 (not significant). FPB: 248.5 and 467.5. Second VLM (gpt-6-luna): 81.6, 37.3 and 225.6 \\
\addlinespace
Labeled photographs of other foods or trays & Fine-tuned with those photographs & ASA24: 42.0 against 72.9 for Qwen. FPB: 161.7 against 248.5 \\
\addlinespace
The food's identity and one labeled photograph of it & Fine-tuned with per-food correction & ASA24: 28.3, against 42.6 uncorrected and 48.0 from the labeled photograph and food area alone \\
\addlinespace
Training data from the same camera and menu & Fine-tuned & Nutrition5k validation: 24.8 against 60.4 for Qwen \\
\bottomrule
\end{tabular}
\end{table*}

\paragraph{Published Nutrition5k results} Table~\ref{tab:context} places our Nutrition5k results beside published ones, for context only. Our validation set is 502 of the 507 dishes in Nutrition5k's official test split, but our model choices were made on it, so its errors are optimistic by an amount we did not estimate. On it, the fine-tuned DINOv2 (24.8\g) lies between the re-implemented Swin-Nutrition (27.5\g) and DPF-Nutrition (21.2\g), with fewer training dishes and without a depth-estimation network. None of the published methods could be tested on our external data, because no usable code or trained weights were available.

\begin{table*}[t]
\centering
\caption{Published Nutrition5k food-mass results, for context only; rows are not directly comparable.}
\label{tab:context}
\small
\begin{tabular}{L{5.6cm}rrL{2.8cm}L{3.0cm}}
\toprule
Study and method & Dishes & MAE (g) & Inputs & Coverage \\
\midrule
DietAI24~\cite{yan2025dietai24} & 1,000 & 45.1 & Image; FNDDS retrieval & 84.2\% successful; failure handling unspecified \\
Open-KNEAD, Gemma-MoE~\cite{coburn2026openknead} & Not stated & 66.7 & Image & Full test split \\
Liao and Li~\cite{liao2026geometry}, sum of per-food estimates & 502 & 36.6 & Image; LLM food detections & Official test split; trained per dataset \\
\midrule
Nutrition5k RGB baseline~\cite{thames2021nutrition5k} & 507 & 40.4 & Image (supervised) & Official test split \\
ViT pretrained on ImageNet-21k~\cite{andrade2025pretraining} & 10\% of plates & 60.6 & Side-video frames (supervised) & Own 90/10 split by plate \\
Swin-Nutrition~\cite{shao2022swin}, as re-implemented in~\cite{han2023dpf} & 507 & 27.5 & Image (supervised) & Official test split; original trained on side-angle frames and reported no grams \\
RGB-D Nutrition~\cite{shao2023rgbd} & 507 & 21.6 & Image + measured depth (supervised) & Official test split \\
DPF-Nutrition~\cite{han2023dpf} & 507 & 21.2 & Image, predicted depth (supervised) & Official test split \\
\midrule
This study, Qwen3-VL-32B single prompt, validation set & 502 & 60.4 & Image only & 100\% \\
This study, kNN residual correction of single-prompt Qwen, validation set & 502 & 47.7 & Image only & 100\% \\
This study, Qwen3-VL-32B two-step, validation set & 502 & 70.5 & Image; closed ingredient list & 100\% \\
This study, fine-tuned DINOv2, validation set & 502 & 24.8 & Image only & 100\% \\
This study, fine-tuned DINOv2, internal test set (other dishes) & 498 & 19.7 & Image only & 100\% \\
\bottomrule
\end{tabular}
\end{table*}

\paragraph{Reporting} We followed the TRIPOD+AI items~\cite{collins2024tripodai} that apply to a model estimating a continuous quantity from an image: data sources and eligibility (Section~\ref{sec:n5ksets}, Table~\ref{tab:datasets}); the definition of the labels; the model's only input, the photograph; handling of unusable outputs (Section~\ref{sec:vlm}); model development (Section~\ref{sec:training} and Appendix~\ref{app:training}); performance measures with clustered confidence intervals (Section~\ref{sec:stats}); separation of internal and external evaluation; and the limitations. Items on clinical outcomes, participants and implementation do not apply.

\end{document}